\RequirePackage{amsmath, amssymb, accents}  

\documentclass{llncs}

\usepackage{bbm}  
\newcommand{\ubar}[1]{\underaccent{\bar}{#1}}

\usepackage[all]{xy}

\usepackage[pdftex]{graphicx}

\usepackage{epstopdf}
\usepackage{epsfig}

\usepackage{subfigure}

\usepackage[numbers,sort&compress]{natbib} 

\usepackage{url}

\usepackage{xspace}
\newcommand{\sync}{SE-Sync\xspace}

\newcommand{\titlestring}{A Certifiably Correct Algorithm for Synchronization over the Special Euclidean Group}

\usepackage{algorithm}
\usepackage{algpseudocode}
\makeatletter
\let\OldStatex\Statex
\renewcommand{\Statex}[1][0]{%
  \setlength\@tempdima{\algorithmicindent}%
  \OldStatex\hskip\dimexpr#1\@tempdima\relax}
\makeatother

\algnewcommand\algorithmicinput{\textbf{Input:}}
\algnewcommand\Input{\item[\algorithmicinput]}
\algnewcommand\algorithmicoutput{\textbf{Output:}}
\algnewcommand\Output{\item[\algorithmicoutput]}

\newcommand{\R}{\mathbb{R}}

\DeclareMathOperator{\Sym}{Sym}

\DeclareMathOperator{\tr}{tr}

\DeclareMathOperator{\rank}{rank}
\DeclareMathOperator{\Diag}{Diag}
\DeclareMathOperator \vect{vec}

\def \transpose{^\mathsf{T}}
\def \inv{^{-1}}
\def \tinv{^{-\mathsf{T}}}
\def \pinv {^\dagger}

\DeclareMathOperator{\Stiefel}{St}

\DeclareMathOperator{\grad}{grad}
\DeclareMathOperator{\Hess}{Hess}
\DeclareMathOperator{\proj}{Proj}

\DeclareMathOperator{\Orthogonal}{O}
\DeclareMathOperator{\SO}{SO}
\newcommand{\so}{\mathfrak{so}}
\DeclareMathOperator{\SE}{SE}

\newcommand{\st}{\textnormal{s.t.\; }}

\DeclareMathOperator{\Gaussian}{\mathcal{N}}  
\DeclareMathOperator{\Langevin}{Langevin}  

\def \Nodes {\mathcal{V}}  
\def \Edges {\mathcal{E}}  
\newcommand{\directed}[1]{\overrightarrow{#1}}  
\def \dEdges{\directed{\Edges}}  

\def \edge{\lbrace i,j \rbrace}  
\def \dedge{(i,j)}  

\def\Lap{L}  

\def\incMat{A}  
\def\redIncMat{\ubar{A}}  

\def \cycProjMatrix{\Pi}

\def \CSAIL{Computer Science and Artificial Intelligence Laboratory}
\def \LIDS{Laboratory for Information and Decision Systems}
\def \MathDep{Department of Mathematics}
\def \MIT{Massachusetts Institute of Technology}
\def \MITaddr{Cambridge, MA 02139, USA}

\def \CourInst{Courant Institute of Mathematical Sciences}
\def \CDS{Center for Data Science}
\def \NYU{New York University}
\def \NYUaddr{New York, NY 10012, USA}

\title{\titlestring}
\titlerunning{\titlestring }

\author{David M.\ Rosen\thanks{Corresponding author.  Email: dmrosen@mit.edu}\inst{1} \and Luca Carlone\inst{2} \and Afonso S.\ Bandeira\inst{3} \and \\ John J.\ Leonard\inst{1} }
\institute{\CSAIL, \MIT, \MITaddr \and \LIDS, \MIT, \MITaddr \and \MathDep {} and \CDS, \CourInst, \NYU, \NYUaddr}

\DeclareMathOperator{\BDiag}{BlockDiag}
\DeclareMathOperator{\SymBlockDiag}{SymBlockDiag}

\def \pose{x}
\def \tran{t}
\def \rot{R}

\newcommand{\true}[1]{\ubar{#1}}
\newcommand{\noisy}[1]{\tilde{#1}}
\newcommand{\est}[1]{\hat{#1}}

\def \tpose{\true{\pose}}
\def \ttran{\true{\tran}}
\def \trot{\true{\rot}}

\def \optsym{*}
\def \topt{\tran^\optsym}
\def \Ropt{\rot^\optsym}

\def \Yopt{Y^\optsym}
\def \Zopt{Z^\optsym}

\def \npose{\noisy{\pose}}
\def \ntran{\noisy{\tran}}
\def \nrot{\noisy{\rot}}

\def \tranNoise{\tran^\epsilon}
\def \rotNoise{\rot^\epsilon}

\def \PosesMLE{\est{\pose}_{\textnormal{MLE}}}

\def \RotEst{\est{\rot}}
\def \TranEst{\est{\tran}}
\def \PoseEst{\est{\pose}}

\def \MLEval{p_{\textnormal{MLE}}^*}  
\def \SDPval{p_{\textnormal{SDP}}^*}  
\def \SDPLRval{p_{\textnormal{SDPLR}}^*}

\def \rotsym{\rho}  
\def \transym{\tau}  

\def \TranW{W^{\transym}}  
\def \LapTranW{\Lap(\TranW)}  
  
\def \MeasRotGraph{\noisy{G}^\rotsym}  
\def \MeasRotConLap{\Lap(\MeasRotGraph)}  

\def \nCrossTerms{\noisy{V}}  

\def \nQ{\noisy{Q}}  
\def \tQ{\true{Q}}

\def \tranPrecisions{\Omega}  
\def \nT{\noisy{T}}

\def \exactnessBound{\beta}

\begin{document}

\maketitle

\begin{abstract}
Many important geometric estimation problems naturally take the form of \emph{synchronization over the special Euclidean group}: estimate the values of a set of unknown poses $\pose_1, \dotsc, \pose_n \in \SE(d)$ given noisy measurements of a subset of their pairwise relative transforms $\pose_i\inv \pose_j$.  Examples of this class include the foundational problems of pose-graph simultaneous localization and mapping (SLAM) (in robotics) and camera pose estimation (in computer vision), among others.  This problem is typically formulated as a maximum-likelihood estimation that requires solving a nonconvex nonlinear program, which is computationally intractable in general.  Nevertheless, in this paper we present an algorithm that is able to efficiently recover \emph{certifiably globally optimal} solutions of this estimation problem in a non-adversarial noise regime. The crux of our approach is the development of a semidefinite relaxation of the maximum-likelihood estimation whose minimizer provides the \emph{exact} MLE so long as the magnitude of the noise corrupting the available measurements falls below a certain critical threshold; furthermore, whenever exactness obtains, it is possible to \emph{verify} this fact \emph{a posteriori}, thereby \emph{certifying} the optimality of the recovered estimate. We develop a specialized optimization scheme for solving large-scale instances of this semidefinite relaxation by exploiting its low-rank, geometric, and graph-theoretic structure to reduce it to an equivalent optimization problem on a low-dimensional Riemannian manifold, and then design a Riemannian truncated-Newton trust-region method to solve this reduction efficiently.  We combine this fast optimization approach with a simple rounding procedure to produce our algorithm, \emph{\sync}.  Experimental evaluation on a variety of simulated and real-world pose-graph SLAM datasets shows that \sync is capable of recovering globally optimal solutions when the available measurements are corrupted by noise up to an order of magnitude greater than that typically encountered in robotics applications, and does so at a computational cost that scales comparably with that of direct Newton-type \emph{local} search techniques.

\end{abstract}



\newcommand{\setal}{~\emph{et~al.}\xspace}
\newcommand{\myParagraph}[1]{{\bf #1.}}

\setcounter{footnote}{0}

\section{Introduction}
Over the coming decades, the increasingly widespread adoption of robotic technology in areas such as transportation, medicine, and disaster response has tremendous potential to increase productivity, alleviate suffering, and preserve life.  At the same time, however, these high-impact applications often place autonomous systems in safety- and life-critical roles, where misbehavior or undetected failures can carry dire consequences \cite{Vlasic2016Tesla}.  While empirical evaluation has traditionally been a driving force in the design and implementation of autonomous systems, safety-critical applications such as these demand algorithms that come with clearly-delineated performance guarantees.  This paper presents one such algorithm, \emph{\sync}, an efficient and \emph{certifiably correct} method for solving the fundamental problem of pose estimation.  

Formally, we consider the \emph{synchronization problem} of estimating a collection of unknown poses\footnote{A \emph{pose} is a position and orientation in $d$-dimensional Euclidean space; this is an element of the \emph{special Euclidean group} $\SE(d) \cong \R^d \rtimes \SO(d)$.} based upon a set of relative measurements between them.  This estimation problem lies at the core of many fundamental perceptual problems in robotics; for example, simultaneous localization and mapping (SLAM) \cite{Lu1997Globally} and multi-robot localization \cite{Aragues2011Multi}.  Closely-related formulations also arise in structure from motion \cite{Hartley2013Rotation,Martinec2007Robust} and camera network calibration \cite{Tron2014Distributed} (in computer vision), sensor network localization \cite{Peters2015Sensor}, and cryo-electron microscopy \cite{Singer2011Three}.  These synchronization problems are typically formulated as instances of maximum-likelihood estimation under an assumed probability distribution for the measurement noise.  This formulation is attractive from a theoretical standpoint due to the powerful analytical framework and strong performance guarantees that maximum-likelihood estimation affords \cite{Ferguson1996Course}.  However, this formal rigor comes at the expense of computational tractability, as the 
maximum-likelihood formulation leads to a nonconvex optimization problem that is difficult to solve.  

\myParagraph{Related Work}
In the context of SLAM, the pose synchronization problem  
is commonly solved using iterative 
numerical optimization methods, e.g.\ Gauss-Newton \cite{Lu1997Globally,Kuemmerle2011g20,Kaess2012iSAM2ijrr}, 
gradient descent~\cite{Olson2006Fast,Grisetti2009Nonlinear}, or
trust region methods \cite{Rosen2014RISE}. This approach is particularly attractive because the rapid convergence speed of second-order numerical optimization methods \cite{Nocedal2006Numerical}, together with their ability to exploit the measurement sparsity that typically occurs in naturalistic problem instances~\cite{Dellaert2006Square}, enables these techniques to scale efficiently to large problem sizes while maintaining real-time operation.  However, this computational expedience comes at the expense of \emph{reliability}, as their restriction to \emph{local} search renders these methods vulnerable to convergence to suboptimal critical points, even for relatively small noise levels \cite{Carlone2015Lagrangian,Rosen2015Approximate}.  This observation, together with the fact that suboptimal critical points usually correspond to egregiously wrong trajectory estimates, has motivated two general lines of research. 
The first addresses the \emph{initialization problem}, i.e., 
how to compute a suitable initial guess for iterative refinement; 
examples of this effort are~\cite{Carlone2014Angular,Carlone2015Initialization,Rosen2015Approximate}.
The second aims at a deeper understanding of the global structure of the pose synchronization problem (e.g. number of local minima, convergence basin), see for example \cite{Huang2010SLAM,Huang2012Local,Wang2012Structure}. 

 
\myParagraph{Contribution}
In our previous work~\cite{Carlone2015Duality,Carlone2015Lagrangian,Carlone2016Planar}, we demonstrated that Lagrangian duality provides an effective means of \emph{certifying} the optimality of an in-hand solution for the pose synchronization problem, and could in principle be used to \emph{directly compute} such certifiably optimal solutions by solving a Lagrangian relaxation.  However, this relaxation turns out to be a \emph{semidefinite program} (SDP), and while there do exist mature general-purpose SDP solvers, their high per-iteration computational cost limits their practical utility to instances of this relaxation involving only a few hundred poses,\footnote{This encompasses the most commonly-used interior-point-based SDP software libraries, including SDPA, SeDuMi, SDPT3, CSDP, and DSDP.} whereas real-world pose synchronization problems are typically one to two orders of magnitude larger.

The main contribution of the present paper is the development of a specialized structure-exploiting optimization procedure that is capable of efficiently solving large-scale instances of the semidefinite relaxation in practice. This procedure provides a means of recovering \emph{certifiably globally optimal} solutions of the pose synchronization problem from the semidefinite relaxation within a non-adversarial noise regime in which minimizers of the latter correspond to \emph{exact} solutions of the former.  Our overall pose synchronization method, \emph{\sync}, is thus a \emph{certifiably correct} algorithm \cite{Bandeira2016Probably}, meaning  that it is able to efficiently recover \emph{globally} optimal solutions of generally intractable problems within a restricted range of operation, and \emph{certify} the optimality of the solutions so obtained.  Experimental evaluation on a variety of simulated and real-world pose-graph SLAM datasets shows that our relaxation remains exact when the available measurements are corrupted by noise up to an order of magnitude greater than that typically encountered in application, and that within this regime, \sync is able to recover \emph{globally} optimal solutions from this relaxation at a computational cost comparable with that of direct Newton-type \emph{local} search techniques.

\section{Problem formulation}
\label{Problem_formulation_and_main_results_section}

The $\SE(d)$ synchronization problem consists of estimating the values of a set of $n$ unknown poses $\pose_1, \dotsc, \pose_n  \in\SE(d)$ given noisy measurements of $m$ of their pairwise relative transforms $\pose_{ij} \triangleq  \pose_i^{-1} \pose_j$ ($i \ne j$).  We model the set of available measurements using an undirected graph $G = (\Nodes, \Edges)$ in which the nodes $i \in \Nodes$ are in one-to-one correspondence with the unknown poses $x_i$ and the edges $\edge \in \Edges$ are in one-to-one correspondence with the set of available measurements, and we assume without loss of generality that $G$ is connected.\footnote{If $G$ is not connected, then the problem of estimating the unknown poses $x_1, \dotsc, x_n$ decomposes into a set of independent estimation problems that are in one-to-one correspondence with the connected components of $G$; thus, the general case is always reducible to the case of connected graphs.}   We let $\directed{G} =(\Nodes, \dEdges)$ be a directed graph obtained from $G$ by fixing an orientation for each of its edges, and assume that a noisy measurement $\npose_{ij}$ of each relative pose $\pose_{ij} = (t_{ij}, R_{ij})$  is obtained 
by sampling from the following probabilistic generative model:
\begin{equation}
\label{probabilistic_generative_model_for_noisy_observations}
\begin{aligned}
 \ntran_{ij} &= \ttran_{ij} + \tranNoise_{ij},  &  \tranNoise_{ij} &\sim \Gaussian\left(0, \tau_{ij}^{-1} I_d\right), \\
 \nrot_{ij} &= \trot_{ij} \rotNoise_{ij}, & \quad \rotNoise_{ij} &\sim \Langevin\left(I_d, \kappa_{ij} \right),
\end{aligned} \quad \quad \forall \dedge \in \dEdges.
\end{equation}
Here $\tpose_{ij} = (\ttran_{ij}, \trot_{ij})$ is the true (latent) value of $\pose_{ij}$, $\Gaussian(\mu, \Sigma)$ denotes the standard multivariate Gaussian distribution with mean $\mu \in \R^d$ and covariance $\Sigma \succeq 0$, and $\Langevin(M, \kappa)$ denotes the \emph{isotropic Langevin distribution} on $\SO(d)$ with mode $M \in \SO(d)$ and concentration parameter $\kappa \ge 0$ (this is the distribution on $\SO(d)$ whose probability density function is given by:
\begin{equation}
 \label{isotropic_Langevin_density_definition}
 p(R; M, \kappa) = \frac{1}{c_d(\kappa)} \exp\left(\kappa \tr(M\transpose R) \right)
\end{equation}
with respect to the Haar measure \cite{Chiuso2008Wide}).\footnote{We use a directed graph to model the measurements $\npose_{ij}$ sampled from \eqref{probabilistic_generative_model_for_noisy_observations} because the distribution of the noise corrupting the latent values $\tpose_{ij}$ is not invariant under $\SE(d)$'s group inverse operation.  Consequently, we must keep track of which state $x_i$ was the ``base frame'' for each measurement. }   Finally, we define $\npose_{ji} \triangleq \npose_{ij}^{-1}$, $\kappa_{ji} \triangleq \kappa_{ij}$, $\tau_{ji} \triangleq \tau_{ij}$, and $\nrot_{ji} \triangleq \nrot_{ij}\transpose$ for all $\dedge \in \dEdges$.

Given a set of noisy measurements $\npose_{ij}$ sampled from the generative model \eqref{probabilistic_generative_model_for_noisy_observations}, a straightforward computation shows that a maximum-likelihood estimate $\PosesMLE \in \SE(d)^n$ for the poses $\pose_1, \dotsc, \pose_n$ is obtained as a minimizer of:

\begin{problem}[Maximum-likelihood estimation for $\SE(d)$ synchronization]
 \label{SE3_synchronization_MLE_NLS_problem}
 \begin{equation}
 \label{SE3_synchronization_MLE_NLS_optimization}
\MLEval = \min_{\stackrel{\tran_i \in \R^d}{\rot_i \in \SO(d)}} \sum_{\dedge \in \dEdges} \kappa_{ij} \lVert \rot_j - \rot_i \nrot_{ij} \rVert_F^2 + \tau_{ij} \left \lVert \tran_j - \tran_i - \rot_i \ntran_{ij} \right \rVert_2^2
 \end{equation}
\end{problem}

Unfortunately, Problem \ref{SE3_synchronization_MLE_NLS_problem} is a high-dimensional, nonconvex nonlinear program, and is therefore computationally hard to solve in general. Consequently,  our strategy in this paper will be to \emph{approximate} this problem using a (convex) \emph{semidefinite relaxation} \cite{Vandenberghe1996Semidefinite}, and then exploit this relaxation to search for solutions of the original (hard) problem.  


\section{Forming the semidefinite relaxation}
\label{Forming_the_semidefinite_relaxation_section}
In this section we develop the semidefinite relaxation that we will solve in place of the maximum-likelihood estimation Problem \ref{SE3_synchronization_MLE_NLS_problem}.\footnote{Due to space limitations, we omit all derivations and proofs; please see the extended version of this paper \cite{Rosen2016SESync} for detailed derivations and additional results.}  To that end, our first step will be to rewrite Problem \ref{SE3_synchronization_MLE_NLS_problem} in a simpler and more compact form that emphasizes the structural correspondences between the optimization problem \eqref{SE3_synchronization_MLE_NLS_optimization} and several simple graph-theoretic objects that can be constructed from the set of available measurements $\npose_{ij}$ and the graphs $G$ and $\directed{G}$.  

We define the \emph{translational weight graph} $\TranW = (\Nodes, \Edges, \lbrace \tau_{ij} \rbrace)$ to be the weighted undirected graph with node set $\Nodes$, edge set $\Edges$, and edge weights $\tau_{ij}$ for $\edge \in \Edges$.  The Laplacian $\LapTranW$ of $\TranW$ is then:
\begin{equation}
\label{Laplacian_of_translational_weight_graph}
\LapTranW_{ij} = 
\begin{cases}
\sum_{ \lbrace i,k \rbrace  \in \Edges} \tau_{ik}, & i = j, \\
-\tau_{ij}, & \edge \in \Edges, \\
0, & \edge \notin \Edges.
\end{cases}
\end{equation}
Similarly, let $\MeasRotConLap$ denote the \emph{connection Laplacian} for the rotational synchronization problem determined by the measurements $\nrot_{ij}$ and measurement weights $\kappa_{ij}$ for $\dedge \in \dEdges$; this is the symmetric $(d \times d)$-block-structured matrix determined by  (cf.\ \cite{Singer2012Vector,Wang2013Exact}):
 \begin{equation}
 \label{connection_Laplacian_definition}
 \begin{split}
\MeasRotConLap &\in \Sym(dn) \\
\MeasRotConLap_{ij} &\triangleq 
\begin{cases}
\left(\sum_{ \lbrace i,k \rbrace \in \Edges } \kappa_{ik} \right) I_d, & i = j, \\
- \kappa_{ij} \nrot_{ij}, & \edge \in \Edges, \\
0_{d \times d}, & \edge \notin \Edges.
\end{cases}
 \end{split}
\end{equation}
We also define a few matrices constructed from the set of translational observations $\ntran_{ij}$.  We let $\nCrossTerms \in \R^{n \times dn}$ be the $(1 \times d)$-block-structured matrix with $(i,j)$-blocks determined by:
\begin{equation}
\label{cross_term_matrix_definition}
\nCrossTerms_{ij} \triangleq 
\begin{cases}
\sum_{\lbrace k \in \Nodes \mid (j,k) \in \dEdges\rbrace} \tau_{jk} \ntran_{jk}\transpose, & i = j, \\
-\tau_{ji} \ntran_{ji}\transpose, & (j,i) \in \dEdges, \\
0_{1 \times d}, & \textnormal{otherwise},
\end{cases}
\end{equation}
$\nT \in \R^{m \times dn}$ denote the $(1 \times d)$-block-structured matrix with rows and columns indexed by $e \in \dEdges$ and $k \in \Nodes$, respectively, and whose $(e,k)$-block is given by:
\begin{equation}
\label{nT_matrix_definition}
 \nT_{ek} \triangleq
 \begin{cases}
-\ntran_{kj}\transpose, & e = (k,j) \in \dEdges,\\
0_{1 \times d}, & \textnormal{otherwise},
 \end{cases}
\end{equation}
and $\tranPrecisions \triangleq \Diag(\tau_{e_1}, \dotsc, \tau_{e_m}) \in \Sym(m)$  denote the diagonal matrix indexed by the directed edges $e \in \dEdges$, and whose $e$th element gives the precision of the translational measurement $\ntran_{ij}$ corresponding to that edge. Finally, we also aggregate the rotational and translational state estimates into the block matrices $R \triangleq \begin{pmatrix}  R_1 & \dotsb & R_n \end{pmatrix}  \in \SO(d)^n \subset \R^{d \times dn}$ and $\tran \triangleq \begin{pmatrix} t_1 & \dots & t_n \end{pmatrix} \in \R^{dn}$.

With these definitions in hand, let us return to Problem \ref{SE3_synchronization_MLE_NLS_problem}.  We observe that for a \emph{fixed} value of the rotational states $R_1, \dotsc, R_n$, this problem  reduces to the \emph{unconstrained} minimization of a quadratic form in the translational variables $t_1, \dotsc, t_n \in \R^d$, for which we can find a closed-form solution using a generalized Schur complement \cite[Proposition 4.2]{Gallier2010Schur}.  This observation enables us to \emph{analytically eliminate} the translational states from the optimization problem \eqref{SE3_synchronization_MLE_NLS_optimization}, thereby obtaining the simplified but equivalent maximum-likelihood estimation:

\begin{problem}[Simplified maximum-likelihood estimation]
\label{Simplified_maximum_likelihood_estimation_for_SE3_synchronization}
\begin{subequations}
\label{Simplified_maximum_likelihood_estimation_for_SE3_synchronization_optimization_problem}
\begin{equation}
\MLEval = \min_{\rot \in \SO(d)^n} \tr(\nQ \rot\transpose\rot) 
\end{equation}
\begin{equation}
\label{Q_quadratic_form_definition}
\nQ = \MeasRotConLap + \nT\transpose \Omega^{\frac{1}{2}} \cycProjMatrix \Omega^{\frac{1}{2}} \nT.
\end{equation}
\end{subequations}
\end{problem}
Furthermore, given any minimizer $\Ropt$ of Problem \ref{Simplified_maximum_likelihood_estimation_for_SE3_synchronization}, we can recover a corresponding optimal value $\topt$ for $\tran$ according to:
\begin{equation}
 \label{minimizing_value_of_t_from_minimizing_value_of_R}
 \topt = - \vect\left( \Ropt \nCrossTerms\transpose \LapTranW\pinv \right).
\end{equation}
\noindent In \eqref{Q_quadratic_form_definition} $\cycProjMatrix \in \R^{m \times m}$ is the matrix of the orthogonal projection operator onto $\ker(\incMat(\directed{G}) \Omega^{\frac{1}{2}}) \subseteq \R^{m \times m}$, where $\incMat(\directed{G}) \in \R^{n \times m}$ is the \emph{incidence matrix} of the directed graph $\directed{G}$.  Although $\cycProjMatrix$ is generically dense, by exploiting the fact that it is derived from a sparse graph, we have been able to show that it admits the sparse decomposition:
\begin{equation}
 \label{decomposition_for_orthogonal_projection_operator_into_sparse_matrices}
\cycProjMatrix = I_m - \Omega^{\frac{1}{2}}\redIncMat(\directed{G})\transpose L\tinv L\inv \redIncMat(\directed{G}) \Omega^{\frac{1}{2}}
\end{equation}
where $\redIncMat(\directed{G}) \Omega^{\frac{1}{2}} = LQ_1$ is a thin LQ decomposition of $\redIncMat(\directed{G}) \Omega^{\frac{1}{2}}$ and $\redIncMat(\directed{G})$ is the \emph{reduced incidence matrix} of $\directed{G}$. Note that expression \eqref{decomposition_for_orthogonal_projection_operator_into_sparse_matrices} requires only the sparse lower-triangular factor $L$, which can be easily and efficiently obtained in practice.  Decomposition \eqref{decomposition_for_orthogonal_projection_operator_into_sparse_matrices} will play a critical role in the implementation of our efficient optimization.

Now we derive the semidefinite relaxation of Problem \ref{Simplified_maximum_likelihood_estimation_for_SE3_synchronization} that we will solve in practice, exploiting the simplified form \eqref{Simplified_maximum_likelihood_estimation_for_SE3_synchronization_optimization_problem}.  We begin by relaxing the condition $\rot \in \SO(d)^n$ to $\rot \in \Orthogonal(d)^n$. The advantage of this latter version is that $\Orthogonal(d)$ is defined by a set of (quadratic) orthonormality constraints,  so the orthogonally-relaxed version of Problem \ref{Simplified_maximum_likelihood_estimation_for_SE3_synchronization} is a \emph{quadratically constrained quadratic program}; consequently, its Lagrangian dual relaxation is a \emph{semidefinite program} \cite{Luo2010Semidefinite}:

\begin{problem}[Semidefinite relaxation for $\SE(d)$ synchronization]
\label{dual_semidefinite_relaxation_for_SE3_synchronization_problem}
\begin{equation}
\label{dual_semidefinite_relaxation_for_SE3_synchronization_optimization}
\begin{split}
&\SDPval = \min_{0 \preceq Z \in \Sym(dn)} \tr(\nQ Z) \\
\st &\BDiag_{d\times d}(Z) = \Diag(I_d, \dotsc, I_d)
\end{split}
\end{equation}
\end{problem}

At this point it is instructive to compare the  semidefinite relaxation \eqref{dual_semidefinite_relaxation_for_SE3_synchronization_optimization} with the simplified maximum-likelihood estimation \eqref{Simplified_maximum_likelihood_estimation_for_SE3_synchronization_optimization_problem}.  For any $\rot \in \SO(d)^n$, the product $Z =\rot \transpose \rot$ is positive semidefinite and has identity matrices along its $(d \times d)$-block-diagonal, and so is a feasible point of \eqref{dual_semidefinite_relaxation_for_SE3_synchronization_optimization}; in other words, \eqref{dual_semidefinite_relaxation_for_SE3_synchronization_optimization} can be regarded as a relaxation of the maximum-likelihood estimation obtained by \emph{expanding \eqref{Simplified_maximum_likelihood_estimation_for_SE3_synchronization_optimization_problem}'s feasible set}.  This immediately implies that $\SDPval \le \MLEval$.  Furthermore, if it happens that a minimizer $\Zopt$ of Problem \ref{dual_semidefinite_relaxation_for_SE3_synchronization_problem} admits a decomposition of the form $\Zopt = {\Ropt}\transpose \Ropt$ for some $\Ropt \in \SO(d)^n$, then it is straightforward to verify that this $\Ropt$ is also a minimizer of Problem \ref{Simplified_maximum_likelihood_estimation_for_SE3_synchronization}, and so provides a \emph{globally} optimal solution of the maximum-likelihood estimation Problem \ref{SE3_synchronization_MLE_NLS_problem} via \eqref{minimizing_value_of_t_from_minimizing_value_of_R}.  The crucial fact that justifies our interest in Problem \ref{dual_semidefinite_relaxation_for_SE3_synchronization_problem} is that (as demonstrated empirically in our prior work  \cite{Carlone2015Lagrangian} and investigated in a simpler setting in \cite{Bandeira2016Tightness}) this problem has a \emph{unique} minimizer of just this form so long as the noise corrupting the available measurements $\npose_{ij}$ is not too large.  More precisely, we prove:\footnote{Please see the extended version of this paper \cite{Rosen2016SESync} for the proof.}

\begin{proposition}
\label{A_sufficient_condition_for_exact_recovery_prop}
Let $\tQ$ be the matrix of the form \eqref{Q_quadratic_form_definition} constructed using the true \emph{(}latent\emph{)} relative transforms $\tpose_{ij}$ in \eqref{probabilistic_generative_model_for_noisy_observations}.  There exists a constant $\exactnessBound \triangleq \exactnessBound(\tQ) > 0$ \emph{(}depending upon $\tQ$\emph{)} such that, if $\lVert \nQ - \tQ \rVert_2 < \exactnessBound$, then:
\begin{enumerate}
 \item [$(i)$]  The semidefinite relaxation Problem \ref{dual_semidefinite_relaxation_for_SE3_synchronization_problem} has a unique solution $\Zopt$, and
 \item [$(ii)$] $\Zopt = {\Ropt}\transpose \Ropt$, where $\Ropt \in \SO(d)^n$ is a minimizer of the maximum-likelihood estimation Problem \ref{Simplified_maximum_likelihood_estimation_for_SE3_synchronization}.
\end{enumerate}
\end{proposition}
\section{The \sync \ algorithm}


\subsection{Solving the semidefinite relaxation}
\label{optimization_approach_subsection}

 As a semidefinite program, Problem \ref{dual_semidefinite_relaxation_for_SE3_synchronization_problem} can in principle be solved in polynomial-time using interior-point methods \cite{Vandenberghe1996Semidefinite,Todd2001Semidefinite}.  In practice, however, the high computational cost of general-purpose semidefinite programming algorithms prevents these methods from scaling effectively to problems in which the dimension of the decision variable $Z$ is greater than a few thousand \cite{Todd2001Semidefinite}.  Unfortunately, typical instances of  Problem \ref{dual_semidefinite_relaxation_for_SE3_synchronization_problem} that are encountered in  robotics and computer vision applications are one to two orders of magnitude larger than this maximum effective problem size, and are therefore well beyond the reach of these general-purpose methods.  Consequently, in this subsection we develop a specialized optimization  procedure for solving large-scale instances of Problem \ref{dual_semidefinite_relaxation_for_SE3_synchronization_problem} efficiently. 

\subsubsection{Simplifying Problem \ref{dual_semidefinite_relaxation_for_SE3_synchronization_problem}}
\label{Simplifying_dual_semidefinite_relaxation_section}

The dominant computational cost when applying general-purpose semidefinite programming methods to solve Problem \ref{dual_semidefinite_relaxation_for_SE3_synchronization_problem} is the need to store and manipulate expressions involving the (large, dense) matrix variable $Z$.  However, in the typical case that exactness holds, we know that the actual \emph{solution} $\Zopt$ of Problem \ref{dual_semidefinite_relaxation_for_SE3_synchronization_problem} that we seek has a very concise description in the factored form $\Zopt = {\Ropt}\transpose \Ropt$ for $\Ropt \in \SO(d)^n$.  More generally, even in those cases where exactness fails, minimizers $\Zopt$ of Problem \ref{dual_semidefinite_relaxation_for_SE3_synchronization_problem} generically have a rank $r$ not much greater than $d$, and therefore admit a symmetric rank decomposition $\Zopt = {\Yopt}\transpose {\Yopt}$ for $\Yopt \in \R^{r \times dn}$ with $r \ll dn$. 

In a pair of papers, \citet{Burer2003Nonlinear,Burer2005Local} proposed an elegant general approach to exploit the fact that large-scale semidefinite programs often admit such low-rank solutions: simply replace every instance of the decision variable $Z$ with a rank-$r$ product of the form $Y\transpose Y$ to produce a \emph{rank-restricted} version of the original problem.  This substitution has the two-fold effect of (i) dramatically reducing the size of the search space and (ii) rendering the positive semidefiniteness constraint \emph{redundant}, since $Y\transpose Y \succeq 0$ for \emph{any} choice of $Y$.  The resulting rank-restricted form of the problem is thus a low-dimensional \emph{nonlinear} program, rather than a \emph{semidefinite} program.  

Furthermore, following \citet{Boumal2015Riemannian} we observe that after replacing $Z$ in Problem \ref{dual_semidefinite_relaxation_for_SE3_synchronization_problem} with $Y\transpose Y$ for $Y = \begin{pmatrix} Y_1 & \dotsb & Y_n  \end{pmatrix} \in \R^{r \times dn}$, the block-diagonal constraints in our specific problem of interest \eqref{dual_semidefinite_relaxation_for_SE3_synchronization_optimization} are equivalent to $Y_i\transpose Y_i = I_d$, for all $i \in [n]$, i.e., the columns of each block $Y_i \in \R^{r \times d}$ form an \emph{orthonormal frame}. In general, the set of all orthonormal $k$-frames in $\R^p$ ($k \le n$):
\begin{equation}
 \label{Stiefel_manifold_definition}
 \Stiefel(k, p) \triangleq \left \lbrace Y \in \R^{p \times k} \mid Y\transpose Y = I_k \right \rbrace
\end{equation}
forms a smooth compact matrix manifold, called the \emph{Stiefel manifold}, which can be equipped with a Riemannian metric induced by its embedding into the ambient space $\R^{p \times k}$ \cite[Sec.\ 3.3.2]{Absil2009Optimization}.  Together, these observations enable us to reduce Problem \ref{dual_semidefinite_relaxation_for_SE3_synchronization_problem} to an equivalent \emph{unconstrained} Riemannian optimization problem defined on a product of Stiefel manifolds:

\begin{problem}[Rank-restricted semidefinite relaxation, Riemannian optimization form]
\label{rank_restricted_semidefinite_relaxation_Riemannian_optimization_form_problem}
 \begin{equation}
 \label{rank_restricted_semidefinite_relaxation_Riemannian_optimization_form}
 \SDPLRval = \min_{Y \in \Stiefel(d, r)^n} \tr(\nQ Y\transpose Y).
 \end{equation}
\end{problem}

\noindent This is the optimization problem that we will actually solve in practice.

\subsubsection{Ensuring optimality}
While the reduction from Problem \ref{dual_semidefinite_relaxation_for_SE3_synchronization_problem} to Problem \ref{rank_restricted_semidefinite_relaxation_Riemannian_optimization_form_problem} dramatically reduces the size of the optimization problem that needs to be solved, it comes at the expense of (re)introducing the quadratic orthonormality constraints \eqref{Stiefel_manifold_definition}.  It may therefore not be clear whether anything has really been gained by relaxing Problem \ref{Simplified_maximum_likelihood_estimation_for_SE3_synchronization} to Problem \ref{rank_restricted_semidefinite_relaxation_Riemannian_optimization_form_problem}, since it appears that we may have simply replaced one difficult nonconvex optimization problem with another.  The following remarkable result (adapted from \citet{Boumal2106Nonconvex}) justifies this approach:

\begin{proposition}[A sufficient condition for global optimality in Problem \ref{rank_restricted_semidefinite_relaxation_Riemannian_optimization_form_problem}]
\label{Global_optima_of_rank_restricted_semidefinite_relaxation_Riemannian_optimization_problem_prop}
 If $Y \in \Stiefel(d, r)^n$ is a \emph{(}row\emph{)} rank-deficient second-order critical point\footnote{That is, a point satisfying $\grad F(Y) = 0$ and $\Hess F(Y) \succeq 0$ (cf.\ \eqref{function_and_derivatives_for_rank_restricted_Riemannian_form_of_semidefinite_relaxation}--\eqref{Riemannian_Hessian_vector_product_expression}).} of Problem \ref{rank_restricted_semidefinite_relaxation_Riemannian_optimization_form_problem}, then $Y$ is a global minimizer of Problem \ref{rank_restricted_semidefinite_relaxation_Riemannian_optimization_form_problem} and $\Zopt = Y\transpose Y$ is a solution of the dual semidefinite relaxation Problem \ref{dual_semidefinite_relaxation_for_SE3_synchronization_problem}. 
 \end{proposition}

%

Proposition  \ref{Global_optima_of_rank_restricted_semidefinite_relaxation_Riemannian_optimization_problem_prop} immediately suggests a procedure for obtaining solutions $\Zopt$ of Problem \ref{dual_semidefinite_relaxation_for_SE3_synchronization_problem} by applying a Riemannian optimization method to search successively higher levels of the rank-restricted hierarchy of relaxations \eqref{rank_restricted_semidefinite_relaxation_Riemannian_optimization_form} until a \emph{rank-deficient} second-order critical point is obtained.\footnote{Note that since \emph{every} $Y \in \Stiefel(d,r)^n$ is row rank-deficient for $r > dn$, this procedure is guaranteed to recover an optimal solution after searching at most $dn + 1$ levels of the hierarchy \eqref{rank_restricted_semidefinite_relaxation_Riemannian_optimization_form}.} This algorithm is referred to as the \emph{Riemannian Staircase} \cite{Boumal2015Riemannian,Boumal2106Nonconvex}.

\subsubsection{A Riemannian optimization method for Problem \ref{rank_restricted_semidefinite_relaxation_Riemannian_optimization_form_problem}}
\label{Solving_Problem_9_subsubsection}
 
 Proposition \ref{Global_optima_of_rank_restricted_semidefinite_relaxation_Riemannian_optimization_problem_prop} provides a means of obtaining \emph{global} minimizers of Problem \ref{dual_semidefinite_relaxation_for_SE3_synchronization_problem} by \emph{locally} searching for second-order critical points of Problem \ref{rank_restricted_semidefinite_relaxation_Riemannian_optimization_form_problem}.  To that end, in this subsection we design a Riemannian optimization method that will enable us to rapidly identify these critical points in practice.
 
Equations \eqref{Q_quadratic_form_definition} and \eqref{decomposition_for_orthogonal_projection_operator_into_sparse_matrices} provide an efficient means of computing \emph{products} with $\nQ$ by performing a sequence of sparse matrix multiplications and sparse triangular solves (i.e., without the need to form the dense matrix $\nQ$ directly).  These operations are sufficient to evaluate the objective appearing in Problem \ref{rank_restricted_semidefinite_relaxation_Riemannian_optimization_form_problem}, as well as the corresponding gradient and Hessian-vector products when considered as a function on the ambient Euclidean space $\R^{r \times dn}$:
\begin{equation}
 \label{function_and_derivatives_for_rank_restricted_Riemannian_form_of_semidefinite_relaxation}
F(Y) \triangleq \tr(\nQ Y\transpose Y), \quad \quad \nabla F(Y) = 2 Y \nQ, \quad \quad \nabla^2 F(Y)[\dot{Y}] = 2\dot{Y} \nQ.
\end{equation}
Furthermore, there are simple relations between the ambient Euclidean gradient and Hessian-vector products in \eqref{function_and_derivatives_for_rank_restricted_Riemannian_form_of_semidefinite_relaxation}  and their corresponding Riemannian counterparts when $F(\cdot)$ is viewed as a function restricted to $\Stiefel(d, r)^n \subset \R^{r \times dn}$.  With reference to the orthogonal projection operator \cite[eq.\ (2.3)]{Edelman1998Geometry}:
\begin{equation}
\label{Stiefel_manifold_orthogonal_projection_operator}
\begin{split}
 \proj_Y &\colon T_Y\left(\R^{r \times dn} \right)\to T_Y\left(\Stiefel(d, r)^n \right) \\
 \proj_Y(X) &= X - Y \SymBlockDiag_{d \times d}(Y\transpose X)
\end{split}
\end{equation}
the Riemannian gradient $\grad F(Y)$ is simply the orthogonal projection of the ambient Euclidean gradient $\nabla F(Y)$ (cf.\ \cite[eq.\ (3.37)]{Absil2009Optimization}): 
\begin{equation}
\label{Riemannian_gradient_expression}
 \grad F(Y) = \proj_Y \nabla F(Y).
\end{equation}
Similarly, the Riemannian Hessian-vector product $\Hess F(Y)[\dot{Y}]$ can be obtained as the orthogonal projection of the ambient directional derivative of the gradient vector field $\grad F(Y)$ in the direction of $\dot{Y}$ (cf.\ \cite[eq.\ (5.15)]{Absil2009Optimization}). A straightforward computation shows that this is given by:
\begin{equation}
\label{Riemannian_Hessian_vector_product_expression}
 \begin{split}
\Hess F(Y)[\dot{Y}] &=  \proj_Y(\nabla^2 F(Y)[\dot{Y}] - \dot{Y} \SymBlockDiag_{d \times d}( Y\transpose \nabla F(Y) ) ).
 \end{split}
\end{equation}

Together, equations \eqref{Q_quadratic_form_definition}, \eqref{decomposition_for_orthogonal_projection_operator_into_sparse_matrices}, and \eqref{function_and_derivatives_for_rank_restricted_Riemannian_form_of_semidefinite_relaxation}--\eqref{Riemannian_Hessian_vector_product_expression} provide an efficient means of computing $F(Y)$, $\grad F(Y)$, and $\Hess F(Y)[\dot{Y}]$.  Consequently, we propose to employ the truncated-Newton \emph{Riemannian Trust-Region} (RTR) method \cite{Absil2007Trust,Boumal2016Global} to efficiently compute high-precision estimates of second-order critical points of Problem \ref{rank_restricted_semidefinite_relaxation_Riemannian_optimization_form_problem} in practice.

\subsection{Rounding the solution}
\label{rounding_procedure_subsection}

In the previous subsection, we described an efficient algorithmic approach for computing minimizers $\Yopt \in \Stiefel(d,r)^n$ of Problem \ref{rank_restricted_semidefinite_relaxation_Riemannian_optimization_form_problem} that correspond to solutions $\Zopt = {\Yopt}\transpose \Yopt$ of Problem \ref{dual_semidefinite_relaxation_for_SE3_synchronization_problem}.  However, our ultimate goal is to extract an optimal solution of Problem \ref{Simplified_maximum_likelihood_estimation_for_SE3_synchronization} from $\Zopt$ whenever exactness holds, and a \emph{feasible approximate solution} otherwise.  In this subsection, we develop a rounding procedure satisfying these criteria.  

To begin, let us consider the (typical) case in which exactness obtains; here ${\Yopt}\transpose \Yopt = \Zopt = {\Ropt}\transpose \Ropt$ for some optimal solution $\Ropt \in \SO(d)^n$ of Problem \ref{Simplified_maximum_likelihood_estimation_for_SE3_synchronization}.  Since $\rank(\Ropt) = d$, this implies that $\rank(\Yopt) = d$ as well.  Consequently, letting $\Yopt = U_d \varXi_d V_d\transpose$ denote a (rank-$d$) thin singular value decomposition of $\Yopt$, and defining $\bar{Y} \triangleq \varXi_d V_d\transpose \in \R^{d \times dn}$, it follows that $\bar{Y}\transpose \bar{Y} = \Zopt = {\Ropt}\transpose \Ropt$.  This last equality implies that the $d \times d$ block-diagonal of $\bar{Y}\transpose \bar{Y}$ satisfies $\bar{Y}_i\transpose \bar{Y}_i = I_d$ for all $i \in [n]$, i.e.\ $\bar{Y} \in \Orthogonal(d)^n$.  Similarly, comparing the elements of the first block rows of $\bar{Y}\transpose \bar{Y}$ and ${\Ropt}\transpose \Ropt$ shows that $\bar{Y}_1 \transpose \bar{Y}_j = \Ropt_1 \Ropt_j$ for all $j \in [n]$.  Left-multiplying this set of equalities by $\bar{Y}_1$ and letting $A = \bar{Y}_1 \Ropt_1$ shows $\bar{Y} = A \Ropt$ for some $A \in \Orthogonal(d).$  Since any product of the form $A \Ropt$ with $A \in \SO(d)$ is \emph{also} an optimal solution of Problem \ref{Simplified_maximum_likelihood_estimation_for_SE3_synchronization} (by gauge symmetry), this shows that $\bar{Y}$  is optimal provided that $\bar{Y} \in \SO(d)$ specifically.  Furthermore, if this is not the case, we can always make it so by left-multiplying $\bar{Y}$ by any orientation-reversing element of $\Orthogonal(d)$, for example $\Diag(1, \dotsc, 1, -1)$.  This argument provides a means of recovering an optimal solution of Problem \ref{Simplified_maximum_likelihood_estimation_for_SE3_synchronization} from $\Yopt$ whenever exactness holds.  Moreover, this procedure straightforwardly generalizes to the case that exactness fails, thereby producing a convenient rounding scheme (Algorithm \ref{Rounding_algorithm}).

\begin{algorithm}[t]
\caption{Rounding procedure for solutions of Problem \ref{rank_restricted_semidefinite_relaxation_Riemannian_optimization_form_problem}}
\label{Rounding_algorithm}
\begin{algorithmic}[1]
\Input An optimal solution $\Yopt \in \Stiefel(d, r)^n$ of Problem \ref{rank_restricted_semidefinite_relaxation_Riemannian_optimization_form_problem}.
\Output A feasible point $\RotEst \in \SO(d)^n$.
\Function{RoundSolution}{$\Yopt$}
\State Compute a rank-$d$ truncated singular value decomposition $U_d \varXi_d V_d\transpose$ for $\Yopt$ \
\Statex[1] and assign $\RotEst \leftarrow \varXi_d V_d\transpose$.
\State Set $N_{+} \leftarrow \lvert \lbrace \RotEst_i \mid \det(\RotEst_i)  >0 \rbrace \rvert$.
\If{$N_{+} < \lceil \frac{n}{2} \rceil$}
\State $\RotEst \leftarrow \Diag(1, \dotsc, 1, -1) \RotEst$.
\EndIf
\For{$i = 1, \dotsc, n$}
\State Set $\RotEst_i \leftarrow$ \Call{NearestRotation}{$\RotEst_i$}.  \Comment{See e.g.\ \cite{Umeyama1991Least}}
\EndFor
\State \Return $\RotEst$
\EndFunction
 \end{algorithmic}
\end{algorithm}

\subsection{The complete algorithm}

Combining the efficient optimization approach of Section \ref{optimization_approach_subsection} with the rounding procedure of Section \ref{rounding_procedure_subsection} produces \emph{\sync} (Algorithm \ref{SE_sync_algorithm}), our proposed algorithm for synchronization over the special Euclidean group.  When applied to an instance of $\SE(d)$ synchronization, \sync\  returns a feasible point $\PoseEst \in \SE(d)^n$ for the maximum-likelihood estimation Problem \ref{SE3_synchronization_MLE_NLS_problem} together with the lower bound $\SDPval \le \MLEval$ on Problem \ref{SE3_synchronization_MLE_NLS_problem}'s optimal value.  Furthermore, in the typical case that Problem \ref{dual_semidefinite_relaxation_for_SE3_synchronization_problem} is exact, the returned estimate $\PoseEst = (\TranEst, \RotEst)$ \emph{attains} this lower bound (i.e.\ $F(\nQ \RotEst\transpose \RotEst) = \SDPval$), which thus serves as a \emph{computational certificate} of $\PoseEst$'s correctness.

\begin{algorithm}[t]
\caption{The \sync \ algorithm}
\label{SE_sync_algorithm}
\begin{algorithmic}[1]
\Input  An initial point $Y \in \Stiefel(d, r_0)^n$, $r_0 \ge d + 1$.
\Output A feasible estimate $\PoseEst\in \SE(d)^n$ for the maximum-likelihood estimation Problem \ref{SE3_synchronization_MLE_NLS_problem}, and a lower bound $\SDPval \le \MLEval$ for Problem \ref{SE3_synchronization_MLE_NLS_problem}'s optimal value.
\Function{\sync}{$Y$}
\State Set $\Yopt \leftarrow \Call{RiemannianStaircase}{Y}$.  \Comment{Solve Problems \ref{dual_semidefinite_relaxation_for_SE3_synchronization_problem} \& \ref{rank_restricted_semidefinite_relaxation_Riemannian_optimization_form_problem}}  
\State Set $\SDPval \leftarrow F(\nQ {\Yopt}\transpose \Yopt)$. \Comment{$\Zopt = {\Yopt}\transpose \Yopt$}
\State Set $\RotEst \leftarrow \Call{RoundSolution}{\Yopt}$.  
\State Recover the optimal translational estimates $\TranEst$ corresponding to $\RotEst$ via \eqref{minimizing_value_of_t_from_minimizing_value_of_R}.

\State Set $\PoseEst \leftarrow (\TranEst, \RotEst)$.
\State \Return $\left \lbrace \SDPval, \PoseEst \right \rbrace$
\EndFunction
 \end{algorithmic}
\end{algorithm}

\newcommand{\e}[1]{\cdot 10^{#1}}
\newcommand{\scenario}[1]{\text{#1}} 
\newcommand{\grid}{\scenario{cube}}
\newcommand{\rim}{\scenario{rim}}
\newcommand{\cubicle}{\scenario{cubicle}}
\newcommand{\sphere}{\scenario{sphere}}
\newcommand{\sphereHard}{\scenario{sphere-a}}
\newcommand{\garage}{\scenario{garage}}
\newcommand{\torus}{\scenario{torus}}
\newcommand{\oneloop}{\scenario{circle}}
\newcommand{\intel}{\scenario{INTEL}}
\newcommand{\bovisa}{\scenario{Bovisa}}
\newcommand{\bov}{\scenario{B25b}}
\newcommand{\fra}{\scenario{FR079}}
\newcommand{\frb}{\scenario{FRH}}
\newcommand{\csail}{\scenario{CSAIL}}
\newcommand{\Ma}{\scenario{M3500}}
\newcommand{\Mb}{\scenario{M10000}}
\newcommand{\ATE}{\scenario{ATE}}
\newcommand{\CVX}{\scenario{CVX}}
\newcommand{\NEOS}{\scenario{NEOS}}
\newcommand{\sdptThree}{\scenario{sdpt3}}
\newcommand{\MOSEK}{\scenario{MOSEK}}
\newcommand{\vertigo}{\scenario{Vertigo}}
\newcommand{\SDPA}{\scenario{SDPA}}

\section{Experimental results}

In this section, we evaluate \sync's performance on a variety of simulated and real-world 3D pose-graph SLAM datasets; specifically, we rerun the experiments considered in our previous work on solution verification \cite{Carlone2015Lagrangian} in order to establish a meaningful baseline with previously published results.  We use a MATLAB implementation of \sync \  that takes advantage of the truncated-Newton RTR \cite{Absil2007Trust} method supplied by  Manopt \cite{Boumal2014Manopt}.  Furthermore, for the purposes of these experiments, we fix $r = 5$ in the Riemannian Staircase,\footnote{We have found empirically that this is sufficient to enable the recovery of an optimal solution whenever exactness holds, and that when exactness fails, the rounded approximate solutions $\RotEst$ obtained by continuing up the staircase are not ultimately appreciably better than those obtained by terminating at $r = 5$.} and initialize \sync \ using a randomly sampled point on $\Stiefel(3, 5)^n$.

As a basis for comparison, we also consider the performance of two Gauss-Newton-based alternatives: one initialized using the common odometric initialization (``\textsf{GN-odom}''), and another using the (state-of-the-art) \emph{chordal} initialization \cite{Carlone2015Initialization,Martinec2007Robust} (``\textsf{GN-chord}'').  Additionally, since the goal of this project is to produce a \emph{certifiably correct} algorithm for $\SE(d)$ synchronization, we also track the cost of applying the \emph{a posteriori} solution verification method \textsf{V2} from our previous work \cite{Carlone2015Lagrangian} to verify the estimates recovered by the Gauss-Newton method with chordal initialization (``\textsf{GN-chord-v2}'').

\subsection{\textsf{Cube} experiments}

\begin{figure}
\center
\subfigure{\includegraphics[width=.35\textwidth]{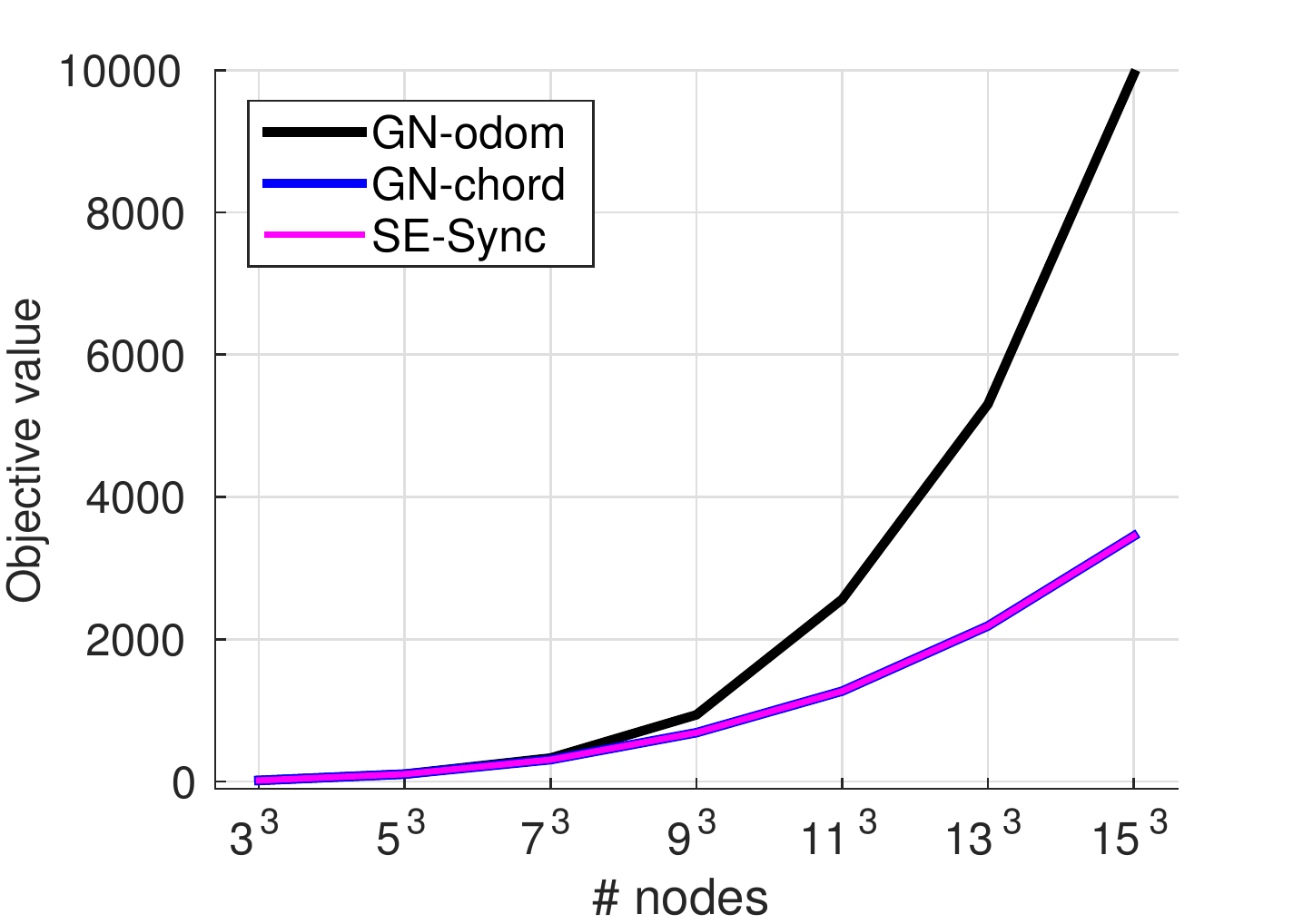}}
\subfigure{\includegraphics[width=.35\textwidth]{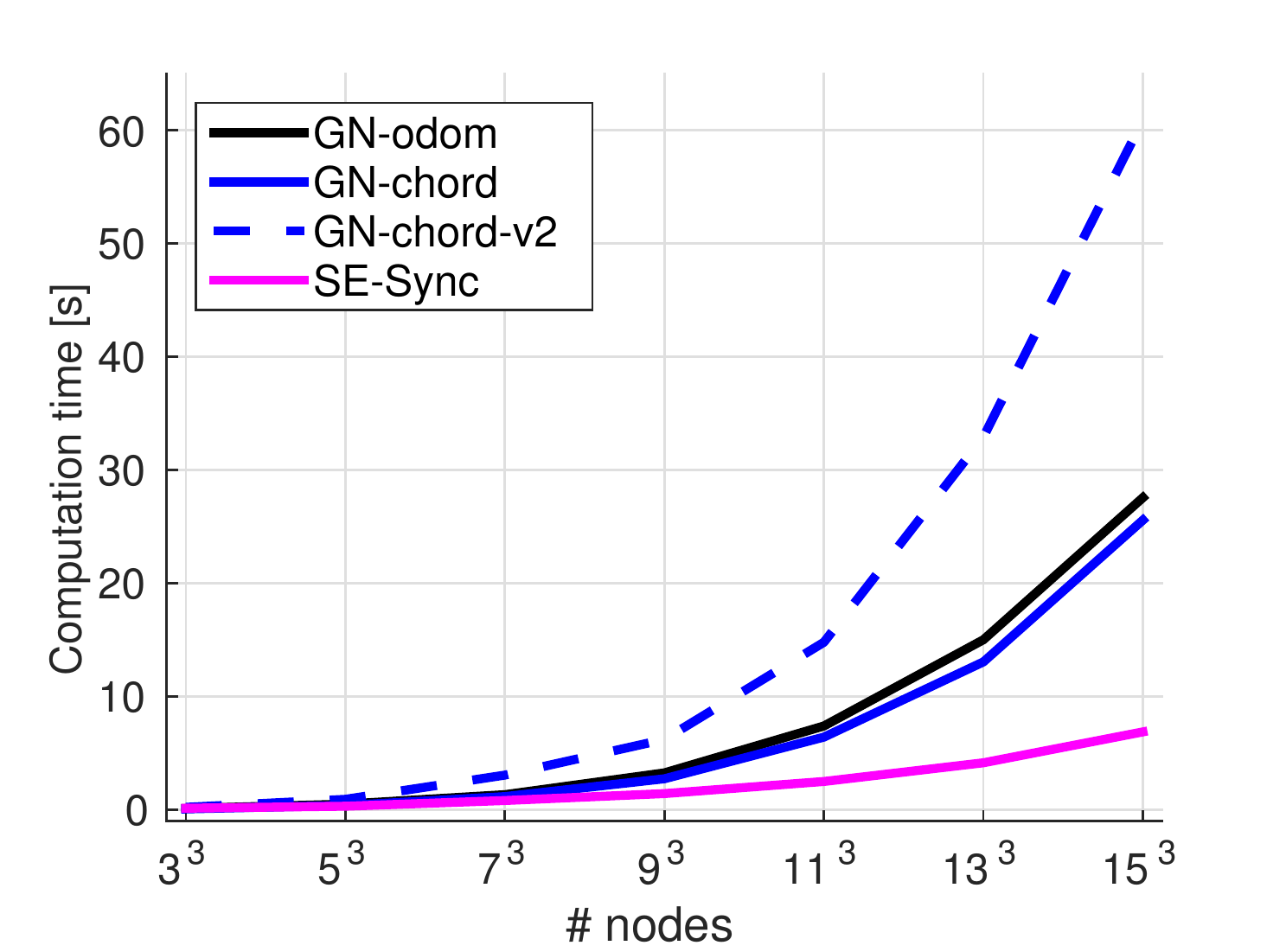}}
\caption{
Varying $s$ while holding $p_{LC} = .1$, $\sigma_R = .1$ rad, $\sigma_T = .5$ m.
}
\label{varying_number_nodes_fig}
\end{figure}

\begin{figure}
\center
\subfigure{\includegraphics[width=.36\textwidth]{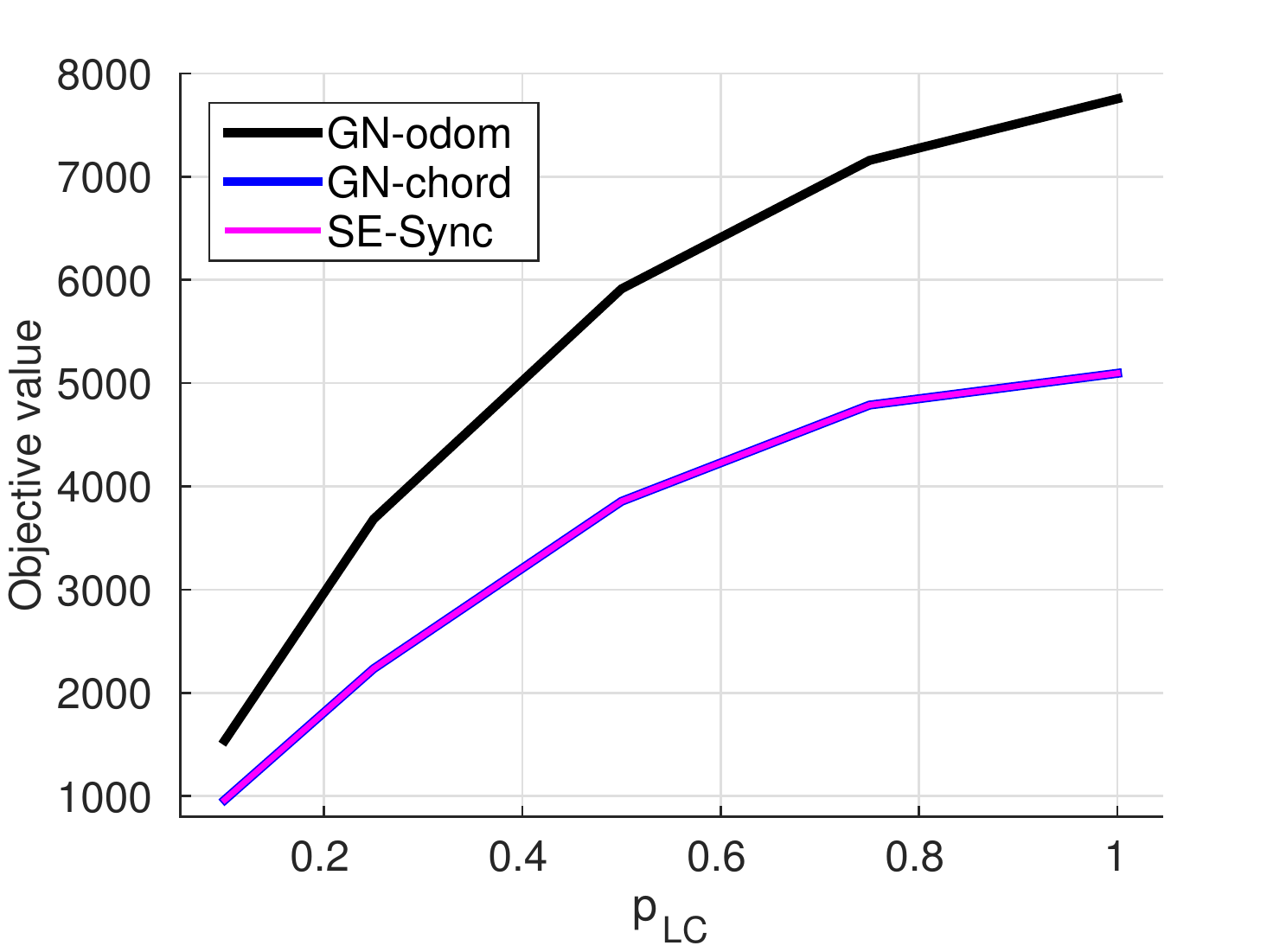}}
\subfigure{\includegraphics[width=.36\textwidth]{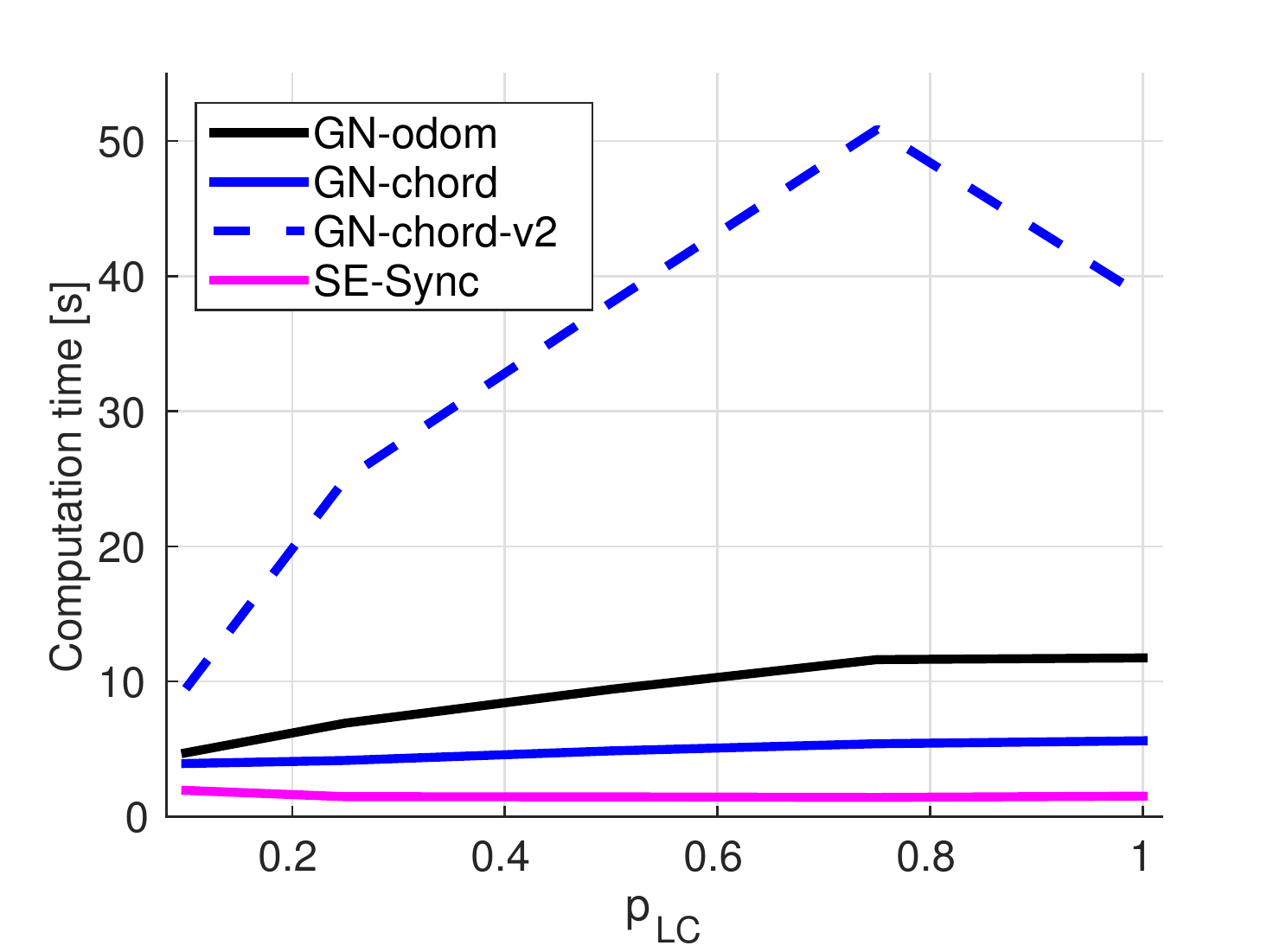} }
\caption{
Varying $p_{LC}$ while holding $s = 10$, $\sigma_R = .1$ rad, $\sigma_T = .5$ m.
}
\label{varying_loop_closure_prob}
\end{figure}

\begin{figure}
\center
\subfigure{\includegraphics[width=.36\textwidth]{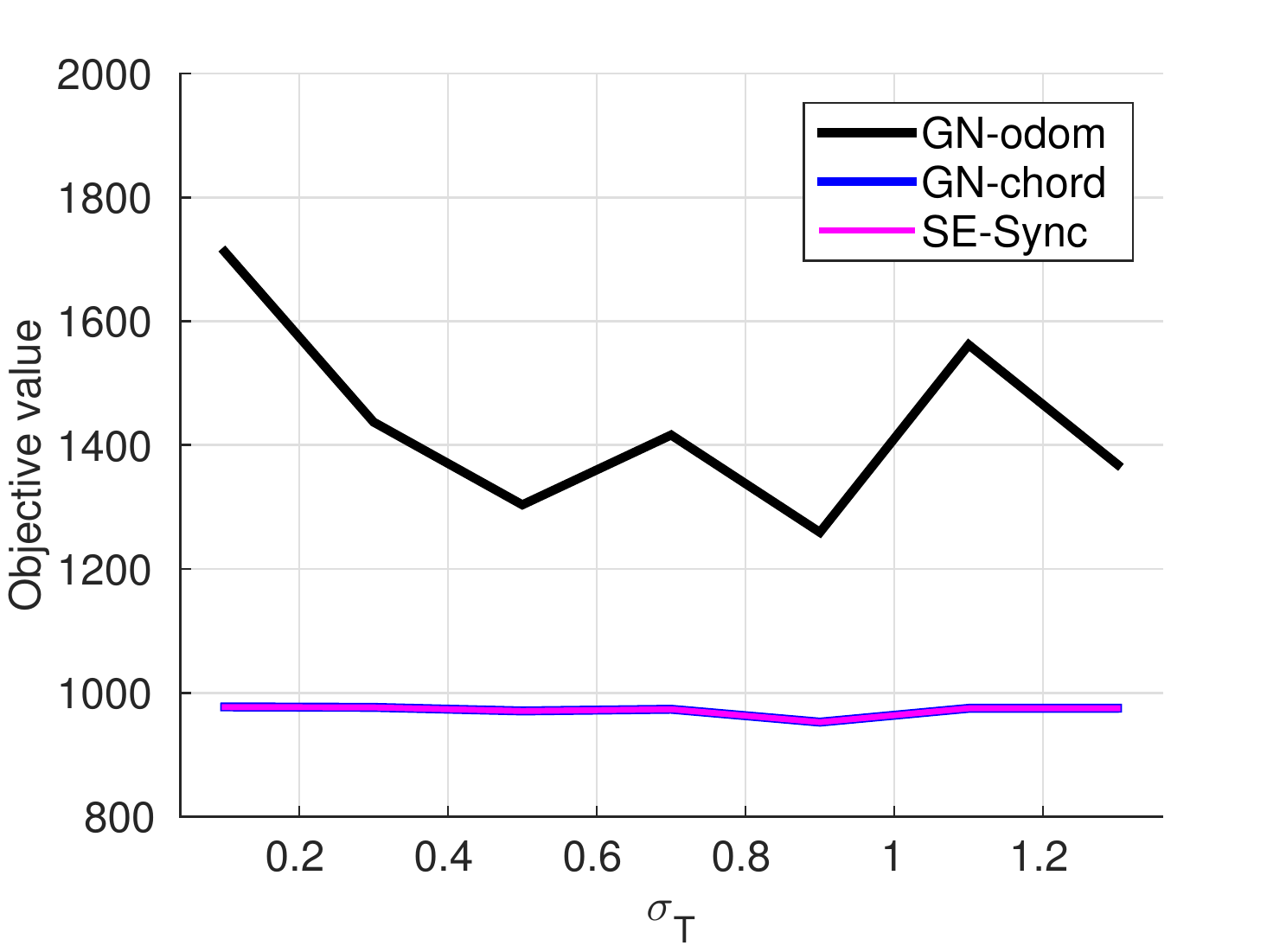}}
\subfigure{\includegraphics[width=.36\textwidth]{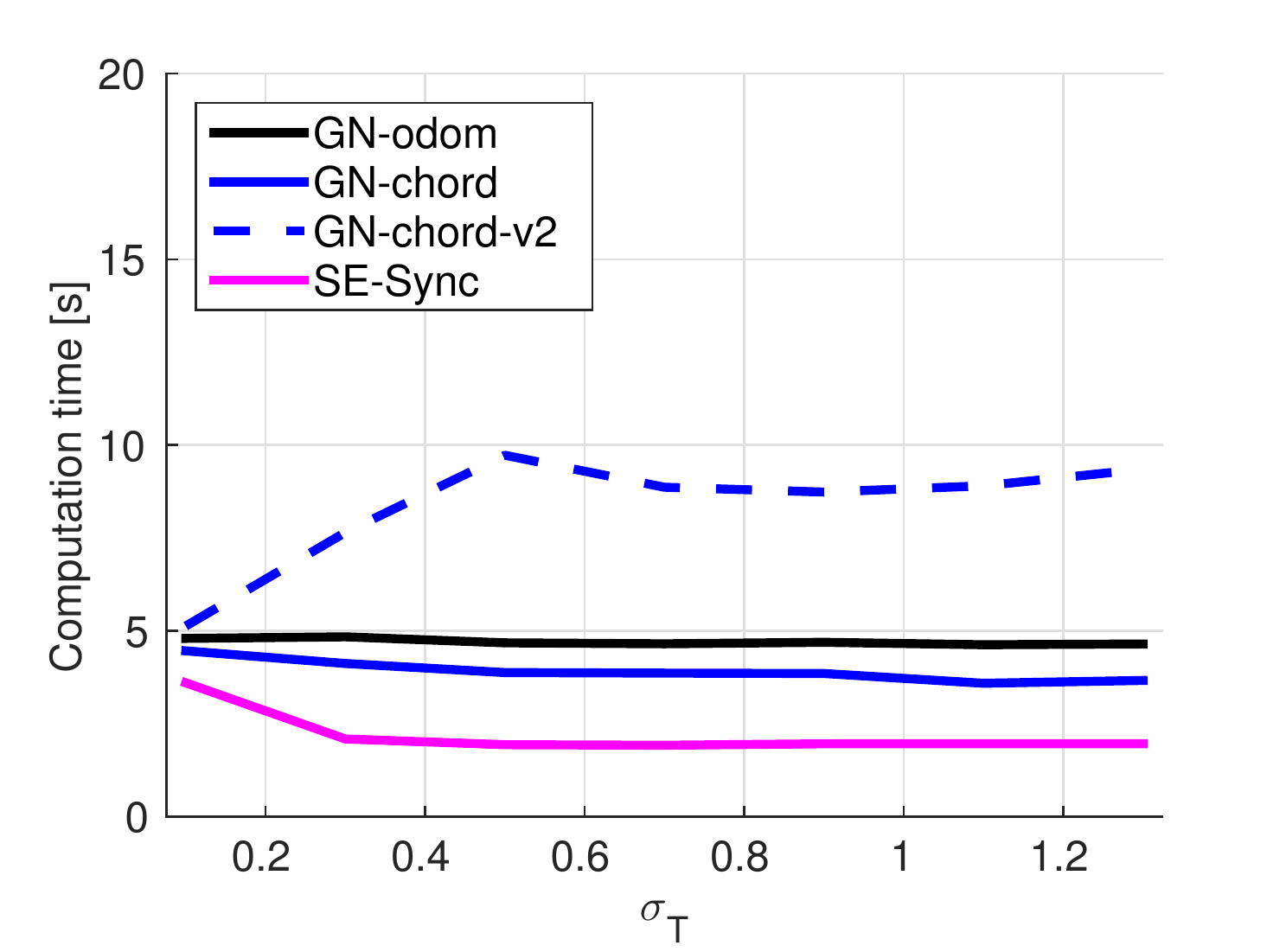} }
\caption{
Varying $\sigma_T$ while holding $s = 10$, $p_{LC} = .1$, $\sigma_R = .1$ rad.
}
\label{varying_translational_noise_level}
\end{figure}

\begin{figure}
\center
\subfigure{\includegraphics[width=.36\textwidth]{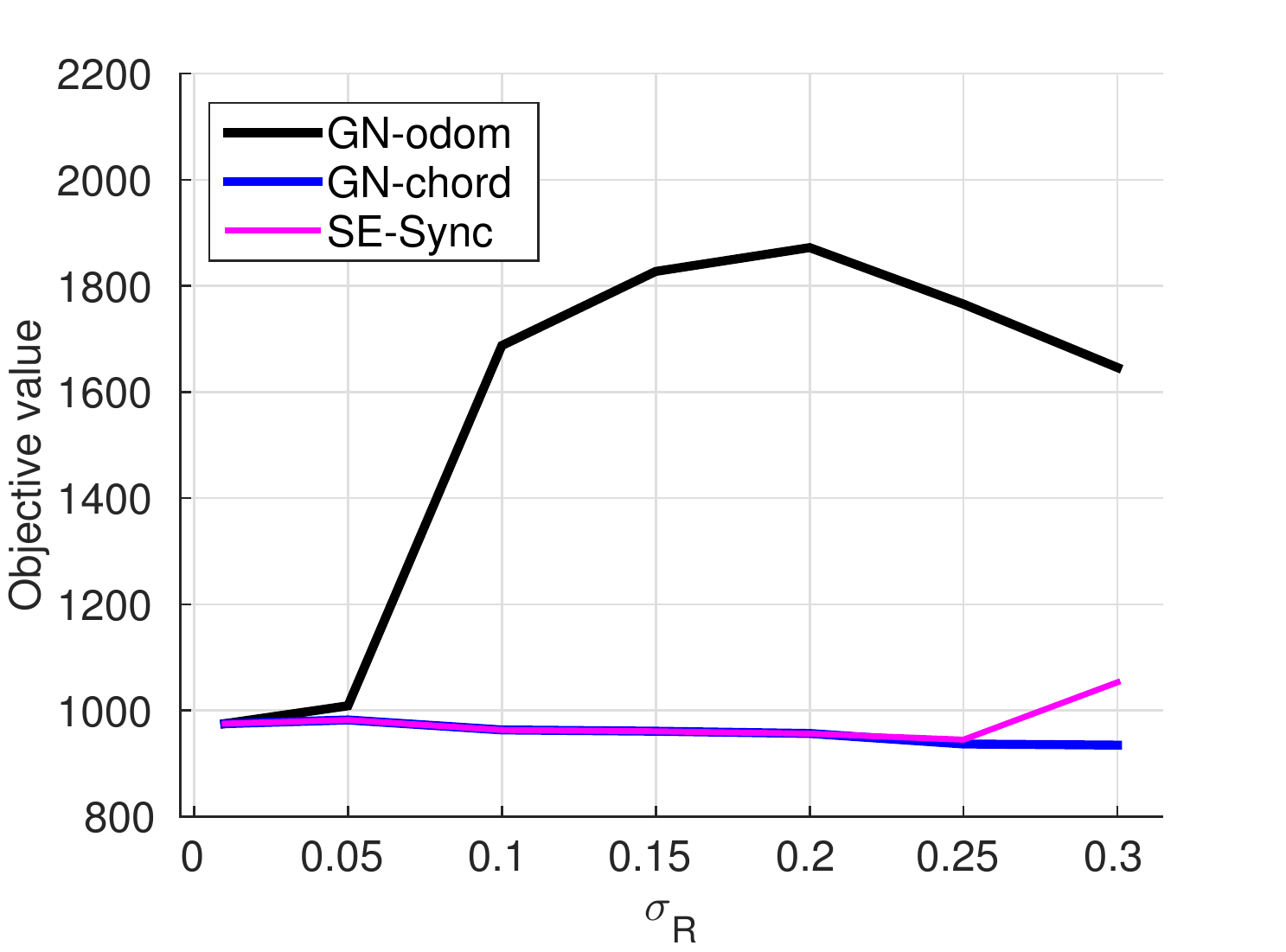}}
\subfigure{\includegraphics[width=.36\textwidth]{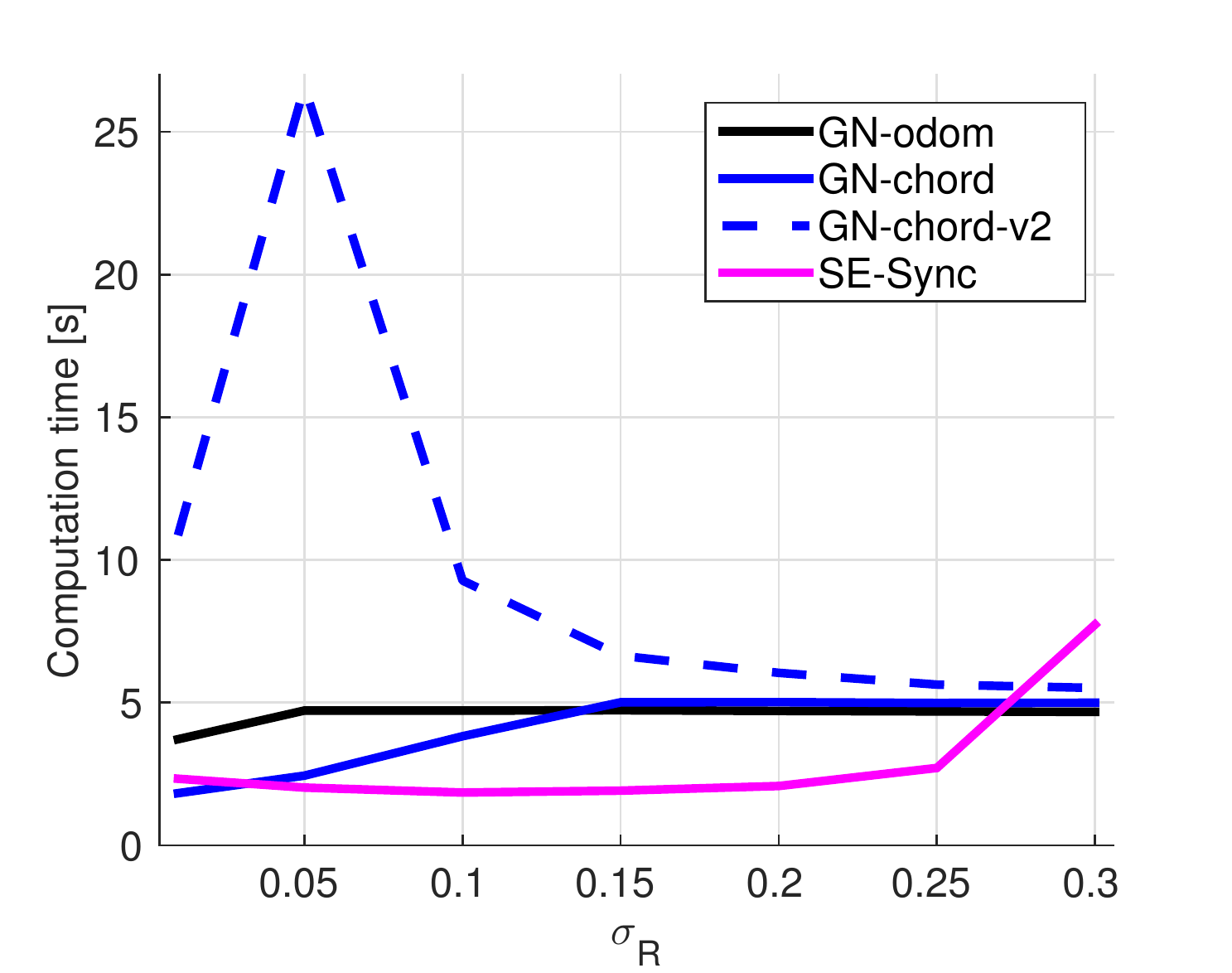} }
\caption{
Varying $\sigma_R$ while holding $s = 10$, $p_{LC} = .1$, $\sigma_T = .5$ m.
}
\label{varying_rotational_noise_level}
\end{figure}

In this first set of experiments, we simulate a robot traveling along a 3D grid world consisting of $s^3$ poses arranged in a cubical lattice.  Random loop closures are added between nearby nonsequential poses on the trajectory with probability $p_{LC}$, and relative measurements are obtained by contaminating translational observations with zero-mean isotropic Gaussian noise with a standard deviation of $\sigma_T$ and rotational observations with the exponential of an element of $\so(3)$ sampled from a mean-zero isotropic Gaussian with standard deviation $\sigma_R$.  Statistics are computed over 30 runs, varying each of the parameters $s$, $p_{LC}$, $\sigma_T$, and $\sigma_R$ individually. Results for these experiments are shown in Figs.\ \ref{varying_number_nodes_fig}--\ref{varying_rotational_noise_level}.
%
%
%
%
%
%


%

Overall, these results indicate that \sync \ is capable of converging to a \emph{certifiably globally optimal} solution from a \emph{random} starting point in a time that is comparable to  (or in these cases, even \emph{faster than}) a standard implementation of a purely \emph{local} search technique initialized with a state-of-the-art method.  This speed differential is particularly pronounced when comparing our (natively certifiably correct) algorithm against the Gauss-Newton method with verification.  Furthermore, this remains true across a wide variety of operationally-relevant conditions.  The one exception to this general rule appears to be in the high-rotational-noise regime (Fig.\ \ref{varying_rotational_noise_level}), where the exactness of the relaxation breaks down (consistent with our earlier findings in \cite{Carlone2015Lagrangian}) and \sync \ is unable to recover a good solution.  Interestingly, the chordal initialization + Gauss-Newton method appears to perform remarkably well in this regime, although here it is no longer possible to certify the correctness of its results (as the certification procedure depends upon the same semidefinite relaxation as does \sync).

\subsection{Large-scale SLAM datasets}

Next, we apply \sync \ to the large-scale SLAM datasets considered in our previous work \cite{Carlone2015Lagrangian}.  We are interested in both the quality of the solutions that \sync \ obtains, as well as its computational speed when applied to challenging large-scale instances of the SLAM problem.  Results are summarized in Table \ref{SLAM_benchmarks_table}.

%
%
%
%
%
%
%
\vspace{-.5\baselineskip}
\begin{table}[h]
\begin{centering}
{\scriptsize 
\resizebox{\textwidth}{!}{  
\begin{tabular}{|  c  ||  c | c || c | c | c || c |c |}
\hline
     & \# Nodes & \# Meas.\ & $f_{GN}$ odom. & $f_{GN}$ init.  & $f_{SE-sync}$   & \sync \ time  & \sync \ optimal?  \\
\hline
\sphere    &$2500$ & $4949$ & $5.802\e{2}$  & $5.760\e{2}$  & $5.759\e{2}$  & $6.95$ & \checkmark  \\ 
\hline 
\sphereHard   &$2200$ & $8647$ & $3.041\e{6}$  & $1.249\e{6}$  & $1.249\e{6}$ & $3.61$ & \checkmark  \\ 
\hline 
\torus   &$5000$ & $9048$ & $2.767\e{4}$  & $1.211\e{4}$  & $1.211\e{4}$  & $24.50$   & \checkmark \\ 
\hline 
\grid   & $8000$ & $22236$ & $2.747\e{5}$  & $4.216\e{4}$  & $4.216\e{4}$  & $132.44$   & \checkmark \\ 
\hline 
\garage   & $1661$& $6275$ & $6.300\e{-1}$  & $6.300\e{-1}$  & $6.299\e{-1}$  & $203.30$  & \checkmark  \\ 
\hline 
\cubicle   & $5750$ & $16869$ & $6.248\e{2}$  & $6.248\e{2}$  & $6.248\e{2}$  & $181.12$ & \checkmark  \\ 
\hline 
\end{tabular}
}
}
\vspace{0.1cm}
\caption{Summary of results for large-scale real-world datasets. }
\label{SLAM_benchmarks_table}
\label{table}
\end{centering}
\vspace{-0.5cm}
\end{table}


These results confirm that the Riemannian optimization scheme developed in Section \ref{optimization_approach_subsection} enables \sync \ to scale effectively to large problem sizes, and that the semidefinite relaxation Problem \ref{dual_semidefinite_relaxation_for_SE3_synchronization_problem} remains exact even in these challenging real-world examples.

\section{Conclusion}

In this paper we developed \sync, a certifiably correct algorithm for synchronization over the special Euclidean group.  This method exploits the same semidefinite relaxation used in our procedure \cite{Carlone2015Lagrangian} to \emph{verify} in-hand optimal solutions, but improves upon this prior work by enabling the efficient \emph{direct computation} of such solutions through the use of a specialized, structure-exploiting low-dimensional Riemannian optimization approach.  Experimental evaluation shows that \sync is capable of recovering \emph{certifiably optimal} solutions of pose-graph SLAM problems in challenging large-scale examples, and does so at a computational cost comparable with that of direct Newton-type \emph{local} search techniques.


\bibliographystyle{plainnat}
\bibliography{references,extra_refs}

\end{document}